\documentclass[sigconf]{acmart}
\usepackage{graphicx}
\usepackage{xcolor}
\usepackage{caption}
\usepackage{cleveref}
\usepackage{enumitem}
\usepackage{booktabs}

\newenvironment{compact_enum}{
  \begin{itemize}[leftmargin=*, itemsep=0pt, parsep=2pt, topsep=2pt]
}{
  \end{itemize}
}

\title{Designing Reward Signals for Portable Query Generation: A Case Study in Industrial Semantic Job Search}

\author{Ping Liu}
\affiliation{
 \institution{LinkedIn Corporation}
 \city{Mountain View}
 \state{CA}
 \country{USA}
}
\email{piliu@linkedin.com}

\author{Qianqi Shen}
\affiliation{
 \institution{LinkedIn Corporation}
 \city{Mountain View}
 \state{CA}
 \country{USA}}
\email{qishen@linkedin.com}

\author{Jianqiang Shen}
\affiliation{
 \institution{LinkedIn Corporation}
 \city{Mountain View}
 \state{CA}
 \country{USA}}
\email{jershen@linkedin.com}

\author{Yunxiang Ren}
\affiliation{
 \institution{LinkedIn Corporation}
 \city{Mountain View}
 \state{CA}
 \country{USA}}
\email{yunren@linkedin.com}

\author{Wenqiong Liu}
\affiliation{
 \institution{LinkedIn Corporation}
 \city{Mountain View}
 \state{CA}
 \country{USA}}
\email{ecliu@linkedin.com}

\author{Chunnan Yao}
\affiliation{
 \institution{LinkedIn Corporation}
 \city{Mountain View}
 \state{CA}
 \country{USA}}
\email{chyao@linkedin.com}

\author{Dan Xu}
\affiliation{
 \institution{LinkedIn Corporation}
 \city{Mountain View}
 \state{CA}
 \country{USA}}
\email{dnxu@linkedin.com}

\author{Baofen Zheng}
\affiliation{
 \institution{LinkedIn Corporation}
 \city{Mountain View}
 \state{CA}
 \country{USA}}
\email{bzheng@linkedin.com}

\author{Rajat Arora}
\affiliation{
 \institution{LinkedIn Corporation}
 \city{Mountain View}
 \state{CA}
 \country{USA}}
\email{rajarora@linkedin.com}

\author{Kevin Kao}
\affiliation{
 \institution{LinkedIn Corporation}
 \city{Mountain View}
 \state{CA}
 \country{USA}}
\email{kkao@linkedin.com}

\author{Wanjun Jiang}
\affiliation{
 \institution{LinkedIn Corporation}
 \city{Mountain View}
 \state{CA}
 \country{USA}}
\email{wanjiang@linkedin.com}

\author{Andrii Soviak}
\affiliation{
 \institution{LinkedIn Corporation}
 \city{Mountain View}
 \state{CA}
 \country{USA}}
\email{asoviak@linkedin.com}

\author{Jingwei Wu}
\affiliation{
 \institution{LinkedIn Corporation}
 \city{Mountain View}
 \state{CA}
 \country{USA}}
\email{jingwu@linkedin.com}

\author{Wenjing Zhang}
\affiliation{
 \institution{LinkedIn Corporation}
 \city{Mountain View}
 \state{CA}
 \country{USA}}
\email{wzhang@linkedin.com}

\copyrightyear{2026}
\acmYear{2026}
\setcopyright{cc}
\setcctype{by}
\acmConference[Agent4IR '26]{AI Agent for Information Retrieva}{August 09--13, 2026}{Jeju Island, Republic of Korea}

\begin{document}

\begin{abstract}
Job-search platforms rely on low-bandwidth query interfaces that often fail to capture the high-dimensional complexity of candidate profiles. We present an end-to-end RLAIF (Reinforcement Learning from AI Feedback) framework to generate \emph{portable} job search queries, terms that abstract away seeker-specific identifiers while preserving generalizable qualifications. This task introduces a highly adversarial reward surface where policy optimization frequently exploits flaws in LLM-as-judge rubrics, resulting in degenerate verbatim-copying behaviors. 

We conducted comprehensive empirical experiments to isolate the impact of optimization mechanics against structured reward engineering. Our results demonstrate that for critic-free optimizers, performance is overwhelmingly dictated by robust reward shaping, rendering the specific choice of algorithm largely immaterial. While critic-free per-rollout baseline methods (RLOO and REINFORCE++) natively resist reward-hacking, the group-relative advantage normalization in GRPO appears uniquely sensitive to spurious reward signals, making it disproportionately susceptible to exploitation. We show that introducing a deterministic, rule-based reward floor to correct for rewards assigned to verbatim copying mitigates this failure mode, resulting in a substantial $+0.147$ quality improvement on a cross-family evaluation judge. Ultimately, we show that the training-time reward model inflates performance gains by $2.4\times$, confirming that the training success is fundamentally dependent on enforcing reward-shaping disciplines rather than selecting alternative optimizers.
\end{abstract}



\maketitle

\section{Introduction}
\label{sec:intro}

The efficiency of a modern job market is fundamentally limited by the \emph{lexical gap} between a candidate’s profile and a recruiter’s job posting. In professional ecosystems like LinkedIn, users navigate this gap through a low-bandwidth interface: the search bar. A job seeker provides a few keywords, and the platform returns a ranked list of job postings. If that query is suboptimal, such as omitting a transferable skill, specifying a role at the wrong level of generality, or failing to translate across language barriers, it acts as a silent filter, limiting opportunities for which the seeker is eminently qualified. 

Query suggestion, the automated transformation of member profile into a concise search query, bridges the gap mentioned above. Effective suggestions are not merely a convenience; they are a precondition for equitable access to the opportunity surface. This is particularly true for newly graduated students, career-transitioning candidates, multilingual workers, and seekers in markets where formal labor vocabulary diverges from lived experience.

Building a production-grade generative suggestion system is not a standard summarization task. It requires navigating a sharp trade-off between two competing properties: \textbf{1. Grounding:} The queries must be strictly derived from the seeker’s actual qualifications to avoid hallucinated career trajectories; \textbf{2. Portability:} The query must abstract away seeker-specific context (e.g., hyper-local geographies, or current employer names) to ensure it retrieves a broad set of market opportunities rather than just a reflection of the seeker's own profile. This tension is further complicated by \textit{heterogeneous sparsity}. Profiles vary from single-sentence headlines to multi-page CVs and often span multiple languages. A model that cannot cleanly ``refuse'' to suggest a query for an uninformative profile creates a liability of fabrication.

While Supervised Fine-Tuning (SFT) provides a baseline, it often fails to capture the nuance of portability and complicated product policy. Reinforcement Learning from AI Feedback (RLAIF) allows us to treat the reward signal as a programmable proxy for human judgment to the product policy. However, shifting to RLAIF moves the core technical challenge from \emph{model architecture} to \emph{reward engineering}. In an industrial context, \textit{Goodhart’s Law}~\citep{goodhart_1975_monetary_management, manheim_2018_goodhart_variants} serves as the primary antagonist. When a policy is optimized against a rubric-prompted LLM judge, it quickly discovers optimization short-circuits that satisfy the rubric's surface features without improving query quality. We observed that the design of the reward signal can be a more potent lever for performance than the choice of RL optimizer. 

This paper provides an end-to-end account of building a rubric-graded RLAIF pipeline on industrial job marketplace data. Our central observation is that \emph{the dominant challenge in industrial RLAIF lies within the reward signal formulation, rather than the choice of optimizer}. Our contributions are as follows:

\begin{compact_enum}
    \item {Formalization of the Portable-Query Problem}, distinguishing it from traditional document expansion and summarization tasks.
    \item A reward-hacking case study characterizing a verbatim-copy failure mode and presenting two concrete architectural mitigations: a deterministic rule-based reward floor ($S_r$) that detects verbatim copying (6-gram profile overlap and lifted date-ranges) to clamp the reward, and a multi-dimensional rubric redesign that explicitly penalizes copying.
    \item Empirical evidence showing that while optimization algorithms (such as GRPO, RLOO, and PPO) cluster closely in baseline performance, the introduction of rule-based reward shaping provides the most significant shift in production quality.
\end{compact_enum}

\section{Related Work}
\label{sec:related-work}

\textit{Query Generation.} Generative query production has evolved from rewriting user input with RL \citep{nogueira_2017_query_reformulation} to expanding documents with predicted queries at indexing or inference time \citep{nogueira_2019_doc2query, bonifacio_2022_inpars, wang_2022_gpl, gao_2022_hyde, wang_2023_query2doc}. While prior talent search work focuses on ranking \citep{ramanath_2018_talent_search_linkedin, geyik_2018_talent_search_linkedin, li_2020_deep_job_understanding}, our system generates a \emph{portable} query from a profile for use across many candidates.

\textit{RL Training for LLMs.} Modern on-policy LLM optimization, traditionally reliant on PPO for RLHF \citep{christiano_2017_human_preferences, stiennon_2020_summarize_human_feedback, ouyang_2022_instructgpt, schulman_2017_ppo}, has increasingly adopted simpler, critic-free architectures that often replace human feedback with AI-driven critiques \citep{shao_2024_deepseekmath_grpo, ahmadian_2024_back_to_basics, hu_2025_reinforce_plus_plus, bai_2022_constitutional_ai, lee_2024_rlaif}. Our work uses a critic-free framework with a prompted rubric grader as its reward signal, rather than a trained reward model.

\textit{Reward Design and Reward Hacking.} Optimizing a policy against a proxy reward risks reward hacking, where the policy exploits the proxy without achieving the intended goal \citep{goodhart_1975_monetary_management, manheim_2018_goodhart_variants, gao_2022_scaling_laws_overoptimization}. Systems can mitigate this with rule-based reward shaping, though AI graders themselves introduce vulnerabilities like sycophancy \citep{ng_1999_reward_shaping, krakovna_2020_specification_gaming, coste_2024_reward_model_ensembles, sharma_2024_sycophancy}. We address this by integrating a multi-dimensional rubric with a deterministic 6-gram copy detector.

\textit{Evaluation of Generative Search Outputs.} LLM-as-a-judge frameworks are now standard for evaluating generative text, replacing brittle reference-based metrics despite known position and length biases \citep{zheng_2023_llm_as_judge, wang_2023_unfair_evaluators, dubois_2024_length_controlled}. Robust validation requires a holistic approach, including dedicated metrics for factual grounding \citep{honovich_2021_q_squared, jacovi_2025_facts_grounding, liang_2023_helm}. Our system thus employs a three-layer evaluation framework with a training-time rubric judge (Layer 1), text-based metrics, and a stricter independent judge ($L_i$).

\section{Problem Setup}
\label{sec:problem-setup}

\subsection{Task: Profile-to-Portable-Query}
\label{subsec:task}

Let $p$ denote a member profile, represented as a structured text document containing the member's headline, current and past experience entries, education history, languages, and (optionally) a summary section. Let $q$ denote a short keyword query, typically $2$--$6$ tokens, that a job seeker would type into a search bar. We seek a policy $\pi_\theta(q \mid p)$ that maps a profile to a query optimized for use by a downstream semantic search engine.

\textit{Portability.}
A query $q$ is \emph{portable} with respect to a profile $p$ if it abstracts the salient transferable qualifications of $p$ --- typically the role, optional seniority, and selective domain --- while omitting tokens that uniquely identify the source member. Concretely, we operationalize portability as the following operational test: would a job seeker's query $q$ retrieve job postings for which a member with profile $p$ is qualified? Anti-examples include verbatim copies of the headline, sub-country location tokens, employer names, date ranges, and other member-specific identifiers. 

Existing generative query systems either (i) generate queries that a \emph{document} could answer to expand the document at indexing time \citep{nogueira_2019_doc2query, bonifacio_2022_inpars, wang_2022_gpl}, (ii) rewrite a user's input query to recall more relevant documents \citep{nogueira_2017_query_reformulation, wang_2023_query2doc, gao_2022_hyde}, or (iii) embed the query side via hypothetical documents \citep{gao_2022_hyde}. None requires \emph{portability} as a first-class property: the queries those systems produce are deliberately specific to the document they were generated from. Our setting reverses this constraint: the query exists to retrieve \emph{other} relevant job candidates, not the source member.

\subsection{Modeling}
\label{subsec:stack}

\textit{Algorithm options.} We evaluate four on-policy advantage estimators on the same infrastructure:

\begin{compact_enum}
  \item \textbf{PPO with critic} \citep{schulman_2017_ppo}: the canonical RLHF baseline. A critic of the same size as the actor is trained alongside. 
  \item \textbf{GRPO}  \citep{shao_2024_deepseekmath_grpo}: samples $K=4$ responses per prompt; the group mean serves as the baseline. No critic.
  \item \textbf{RLOO} \citep{ahmadian_2024_back_to_basics}: samples $K=4$ responses per prompt; per-rollout leave-one-out mean over the group as baseline.
  \item \textbf{REINFORCE++} \citep{hu_2025_reinforce_plus_plus}: global batch-mean baseline with clipped importance ratios.
\end{compact_enum}

All four variants share the same actor initialization (a Qwen3-1.7B model fine-tuned by SFT), the same grader (Qwen3-8B with rubric prompt), the same KL anchor ($\beta = 10^{-3}$, KL loss form), the same data pipeline. The batch size is $48$ per step, the learning rate is $1 \times 10^{-6}$, and the total step budget is $\sim 1{,}600$ steps for one epoch. The only differences across runs are the reward estimator and, where applicable, the rule-based shaping term.

\textit{Reward shaping: $r_{\text{rubric}}$ and $S_r$.}
We use ``reward shaping'' in the classical RL sense \citep{ng_1999_reward_shaping}: any transformation of reward intended to make the learning signal denser, lower-variance, or more aligned with the task objective than the raw signal would be on its own. Our reward signal $r$ on each rollout is composed of two parts:

\begin{compact_enum}
  \item \textbf{$r_{\text{rubric}}$ --- LLM-judge rubric reward.} A pretrained Qwen3-$8$B grader is prompted with the training rubric $R_t$ (\Cref{subsec:rubric-versions}) and a (p, $q_1$) pair, and returns an integer score $N \in \{1, \ldots, 5\}$. We normalize to $r_{\text{rubric}} = (N - 3) / 2 \in [-1, +1]$. The grader is \emph{prompted, not learned} (no reward-model fine-tuning step), so it inherits any blind spots of the underlying LLM and the rubric definition.
  \item \textbf{$S_r$ --- Rule-based deterministic correction.} A small program is applied at reward-parse time, before the grader is invoked: if $q_1$ contains a $6$-gram that appears verbatim in the input profile, \emph{or} a date-range fragment lifted from a profile entry, the reward is clamped to $-1.0$ and the grader is not invoked. $S_r$ is deterministic, cheap, and high-precision; it catches only the two surface signatures the $8$B grader was empirically likely to reward spuriously, leaving the rubric grader responsible for everything else.
\end{compact_enum}

The four-algorithm baseline uses $r_{\text{rubric}}$ alone; GRPO+$S_r$ adds $S_r$ on top. GRPO vs.\ GRPO+$S_r$ contrast isolates the contribution of rule-based shaping on top of the rubric reward, holding the algorithm and the rubric fixed.

\textit{Rubric versions: $R_t$ (training-time) and $R_e$ ($L_i$ evaluation).}
\label{subsec:rubric-versions}
Two rubrics appear in this paper, with disjoint roles:

\begin{compact_enum}
  \item \textbf{$R_t$ --- training rubric.} Five dimensions, scored Pass / Partial / Fail by an LLM judge: D1 Member Understanding; D2 Role, Seniority and Domain; D3 Locale; D4 Portability; D5 Conciseness. The judge prompt embeds the dimension definitions, hard-fail rules, and few-shot examples, and emits a single integer score $1$--$5$ derived from hard rules over the per-dimension Pass/Partial/Fail labels. $R_t$ is the only rubric the policies are trained against (via the Qwen3-$8$B grader); it is also the Layer~1 evaluation rubric.
  \item \textbf{$R_e$ --- evaluation-only rubric.} Adds one dimension on top of $R_t$: D6 Inference Discipline. D6 penalizes over-refusal patterns (e.g.\ emitting \texttt{ROLE MISSING} when the profile actually contains a clear role signal) and over-generalization patterns that the portability and member-understanding dimensions of $R_t$ miss. The hard-fail rule is extended to fail on D4 or D6, and the partial rule extends similarly. $R_e$ is used \emph{only} by $L_i$ (the independent Llama-3.3-$70$B-Instruct grader) at evaluation time. No policy in this paper is trained against $R_e$.
\end{compact_enum}

The asymmetric use is deliberate: training against $R_t$ and evaluating against $R_e$ keeps the independent judge genuinely independent --- a policy can over-fit to $R_t$'s blind spots, and D6 is a new dimension along which that over-fit becomes visible. The Layer-$1$-to-$L_i$ score gap reported in \Cref{sec:experiments-v2.1} is in part an $R_t$-to-$R_e$ rubric gap and in part a Qwen-to-Llama family gap; we do not attempt to disentangle the two in this paper, but the cross-family Llama judge on $R_t$ alone behaves qualitatively similarly (data not shown).

\textit{Reward signal pipeline.}
Each rollout produces a JSON object containing up to three keyword queries; we score only the first query $q_1$ in our experiments. The reward processor (i) parses $q_1$ from the actor's JSON output, (ii) optionally applies $S_r$ which clamps reward to $-1$ on match and short-circuits the grader call, (iii) otherwise invokes the rubric grader on the (profile, $q_1$) pair and emits $r_{\text{rubric}}$. The KL anchor is applied separately by \texttt{verl} \footnote{https://github.com/verl-project/verl} via the actor KL loss against the SFT reference policy.

\textit{Framework.}
We implement on-policy RL on top of \texttt{verl} version $0.7.0$ with \texttt{vLLM} version $0.10.0$ for actor rollouts. Training runs on a single H100 8-GPU node, with $4$ GPUs allocated to the actor rollout pool and $4$ GPUs allocated to the grader pool. The grader is hosted via \texttt{vLLM} in the reward-manager process, with tensor-parallel degree $4$ for the $8$B grader. A custom \textit{Grader Reward Manager} batches rollout prompts into the grader's rubric, parses scores, and emits rewards. Reward is parsed and normalized to $r \in [-1, +1]$. Parse failures emit hard $r = -1$ without expending grader compute.

\section{Experiments}
\label{sec:experiments-v2.1}

To guard against trainer-evaluator inflation, our primary evaluation judge, $L_i$, is an independent, cross-family model (Llama-3.3-70B-Instruct) that grades outputs against the held-out evaluation rubric $R_e$. We sampled $\sim 80$k training rows after a profile-length filter at $300$ characters, and held out $\sim 50$k test rows from different countries. The training-time Layer 1 judge (Qwen3-8B) scores all $\sim 50$k test rows, while the more expensive $L_i$ judge scores a $1{,}000$-row subset; on that subset we report a $95\%$ bootstrap confidence interval, whose half-width of approximately $\pm 0.03$ is our threshold for treating $L_i$ differences as statistically tied.

\begin{table*}[h]
\centering
\small
\caption{Comprehensive Evaluation Leaderboard. Best $L_i$ result in bold; column definitions below the table.}
\label{tab:unified-leaderboard}
\begin{tabular}{l c c c c c c}
\toprule
 & \multicolumn{2}{c}{\textbf{Primary: $L_i$}} & \multicolumn{4}{c}{\textbf{Secondary: Layer 1 Baseline (8B $R_t$)}} \\
\cmidrule(lr){2-3} \cmidrule(lr){4-7}
\textbf{Optimization Model / Algorithm} & \textbf{Mean} & \textbf{$\Delta$ vs SFT} & \textbf{Mean} & \textbf{Tier 1 \%} & \textbf{Tier 5 \%} & \textbf{$q_1$ Chars} \\
\midrule
SFT Framework Initialization (Baseline)  & $+0.595$ & $\phantom{+}0$ & $+0.282$ & $30.5\%$ & $57.0\%$ & $14.1$ \\
GRPO + $S_r$ Linear Correction           & $\mathbf{+0.706}$ & $+0.111$ & $+0.548$ & $19.1\%$ & $73.7\%$ & $17.2$ \\
REINFORCE++ Policy Optimization        & $+0.702$ & $+0.107$ & $+0.524$ & $20.1\%$ & $72.4\%$ & $16.7$ \\
RLOO Leave-One-Out Scaling               & $+0.688$ & $+0.093$ & $+0.525$ & $20.4\%$ & $72.8\%$ & $17.2$ \\
PPO with GAE Critic                    & $+0.612$ & $+0.017$ & $+0.389$ & $26.8\%$ & $64.6\%$ & $18.3$ \\

GRPO without $S_r$ Floor     & $+0.559$ & $-0.036$ & $+0.374$ & $28.4\%$ & $65.1\%$ & $17.1$ \\
GRPO with Simpler Rubric      & $-0.646$ & $-1.241$ & $-0.815$ & $82.7\%$ & $1.8\%$  & $105.9$ \\
\bottomrule
\end{tabular}

\par\smallskip
{\footnotesize\raggedright
$L_i$: independent Llama-3.3-70B on a $1{,}000$-row subset ($95\%$ bootstrap CI; differences within $\pm 0.03$ tied). Layer 1: Qwen3-8B training judge ($R_t$) on all $\sim 50$k rows. Tier 1/5\,\%: fraction in lowest/highest tier; $q_1$ Chars: mean first-query length.\par}
\end{table*}

\paragraph{Two negative controls.}
The bottom two rows of \Cref{tab:unified-leaderboard} are distinct negative controls and should not be conflated:
\begin{compact_enum}
    
  \item \textbf{GRPO (without $S_r$ Floor)} is trained against the current rubric $R_t$ --- the same reward signal as the SOTA cluster --- but without the deterministic $S_r$ correction. It isolates the contribution of rule-based shaping on top of a healthy rubric. The mild $-0.036$ regression below SFT suggests that GRPO's group-standardized advantage normalization may be specifically sensitive to amplifying occasional false-positive signals from the reward model, a documented GRPO bias \cite{liu_2025_understanding_r1_zero}, an effect which $S_r$ mitigates by correcting these rewards.
  \item \textbf{GRPO with simpler rubric} is a variant trained against an simpler pre-$R_t$ rubric that did not penalize verbatim profile copying. It is re-graded under $R_t$ (Layer~1) and $R_e$ ($L_i$) here for comparability. The catastrophic Tier-$1$ density ($82.7\%$ on Layer~1) and $q_1$ length explosion ($105.9$ chars vs.\ $14$--$18$ for healthy policies) reflect a \emph{rubric-design} failure that no choice of optimizer or shaping floor could repair on its own. Both judges identify the failure ($-0.815$ on Layer~1, $-0.646$ on $L_i$), confirming the new rubric stack works as intended.
\end{compact_enum}

\subsection{Empirical Validation of Hypotheses}
\label{subsec:empirical-validation}

\Cref{tab:unified-leaderboard} allows us to evaluate three central hypotheses: (H1) the degree of algorithmic variance between PPO, GRPO, RLOO, and REINFORCE++ once the reward signal is fixed; (H2) the shaping effects of the rule-based $S_r$ floor and its uniformity across algorithms; and (H3) the inflation in quality scores when using the training-time Layer 1 judge versus an independent $L_i$ judge.

\paragraph{H1 (Algorithmic Variance):} Conditioned on the implementation of the $S_r$ shaping constraint within GRPO, the performance delta across all three critic-free paradigms contracts to just $0.018$ score on $L_i$. Because this variation falls below our established statistical significance bound ($\pm 0.03$), we conclude that these optimization approaches achieve equivalent empirical performance bounds. Crucially, the GAE (Generalized Advantage Estimation) -supported PPO configuration lags significantly ($\sim0.09$ units below the critic-free tier), failing to justify its $\sim30\%$ training wall-time computational overhead.

\paragraph{H2 (Deterministic Floor Effects):} Omitting the deterministic floor from the GRPO optimization pipeline (yielding unconstrained GRPO) triggers a systematic drop below the initial baseline SFT capabilities ($-0.036$ $\Delta$). Conversely, adding the rule-based $S_r$ penalty yields an absolute performance improvement of $+0.147$ units on the primary judge. This delta represents the single largest performance shift in the entire empirical matrix, vastly outweighing cross-algorithmic variance. 

The underlying Tier distributions show this effect directly: incorporating the $S_r$ constraint reduces the lowest-Tier degenerate reward density from $28.4\%$ down to $19.1\%$, while shifting peak probability mass toward optimal classifications ($73.7\%$). Intriguingly, both RLOO ($+0.688$) and REINFORCE++ ($+0.702$) successfully converge to the SOTA cluster \emph{without} integrating the structural $S_r$ floor, maintaining verified profile copy-hack frequencies of $0\%$. The necessity of explicit reward shaping ($S_r$) is therefore determined by the choice of advantage estimator, proving essential for group-relative methods while per-rollout-baseline estimators are natively robust to its absence.

\paragraph{H3 (Trainer-Evaluator Inflation Factors):} Localized evaluation against the active training-time judge (Layer 1) reports an apparent performance gain of $+0.265$ above the SFT baseline for the shaped GRPO configuration. However, independent cross-family evaluation ($L_i$) drops this optimization delta to $+0.111$. The active training grader overestimates the magnitude of the policy improvement by a factor of $2.4\times$.

\section{Conclusion}
\label{sec:conclusion}

In this work, we formalized the profile-to-portable-query task, demonstrating that reward signal engineering exerts a significantly greater influence on production quality than algorithmic variance. Under an independent, cross-family validation judge $L_i$, critic-free architectures—including RLOO, REINFORCE++, and a properly shaped GRPO configuration—achieved statistical parity, outperforming the baseline SFT initialization by approximately $+0.10$ units. Conversely, traditional PPO with a GAE critic failed to yield a meaningful quality lift, making its $\sim 30\%$ training overhead unjustifiable.

Our analysis of the interaction between optimization and reward structure suggests that while per-rollout baselines (RLOO and REINFORCE++) natively resist exploitation, GRPO appears more sensitive to the reward model's spurious high rewards for copied text. Its group-standardized advantage can amplify such outliers (consistent with documented biases of GRPO's normalization \cite{liu_2025_understanding_r1_zero}), whereas per-rollout baselines are not. Implementing a deterministic reward-shaping floor ($S_r$) successfully controls this vulnerability by correcting the spurious reward for verbatim copying, yielding a $+0.147$ quality improvement. Finally, we documented a $2.4\times$ performance inflation from the training-time judge compared to the independent evaluation. These outcomes confirm that successful industrial RLAIF deployment depends less on selecting alternative optimizers and more on enforcing rigorous reward-shaping disciplines.

\bibliographystyle{ACM-Reference-Format}
\bibliography{main}

\end{document}